\documentclass{article}
\usepackage{microtype}
\usepackage{graphicx}
\usepackage{subfigure}
\usepackage{booktabs} 
\usepackage{tabularx}

\usepackage{hyperref}

\usepackage[accepted]{icml2023}

\usepackage{amsmath}
\usepackage{dutchcal}
\usepackage{times}
\usepackage{soul}
\usepackage{url}
\usepackage[utf8]{inputenc}
\usepackage[small]{caption}
\usepackage{graphicx}
\usepackage{amsmath}
\usepackage{amsthm}

\usepackage{moreverb,url}

\usepackage[utf8]{inputenc}
\usepackage{dsfont}
\usepackage{graphicx}
\usepackage{float}
\usepackage{algorithm}

\usepackage{booktabs}

\usepackage[title,toc,titletoc,page]{appendix}

\usepackage{tikz}
\usepackage{amsmath}
\usepackage{dsfont}
\usepackage{graphicx}
\usepackage{dblfloatfix}
 \usepackage{amsbsy}
\usepackage{float}

\usepackage{siunitx}
\usepackage{esvect}

\usepackage{xcite}

\usepackage{xr-hyper}
\makeatletter
\newcommand*{\addFileDependency}[1]{
  \typeout{(#1)}
  \@addtofilelist{#1}
  \IfFileExists{#1}{}{\typeout{No file #1.}}
}
\makeatother

\newcommand*{\myexternaldocument}[1]{%
    \externaldocument{#1}%
    \addFileDependency{#1.tex}%
    \addFileDependency{#1.aux}%
}

\myexternaldocument{SI}

\usepackage{bbm}
\usepackage{siunitx}

\newcommand{\vx}{\vv{x}}

\newcommand{\citeay}[1]{\citeauthor{#1} [\citeyear{#1}]}
\newcommand\BibTeX{{\rmfamily B\kern-.05em \textsc{i\kern-.025em b}\kern-.08em
T\kern-.1667em\lower.7ex\hbox{E}\kern-.125emX}}

\icmltitlerunning{Expertise Trees Resolve Knowledge Limitations in Collective Decision-Making }

\begin{document}

\twocolumn[
\icmltitle{Expertise Trees Resolve Knowledge Limitations in Collective Decision-Making}

\icmlsetsymbol{equal}{*}

\begin{icmlauthorlist}
\icmlauthor{Axel Abels}{ulb,vub}
\icmlauthor{Tom Lenaerts}{ulb,vub,chai}
\icmlauthor{Vito Trianni}{rome}
\icmlauthor{Ann Nowé}{vub}
\end{icmlauthorlist}

\icmlaffiliation{ulb}{Machine Learning Group, Universit{\'e} Libre de Bruxelles, Brussels, Belgium}

\icmlaffiliation{vub}{AI Lab, Vrije Universiteit Brussel}

\icmlaffiliation{chai}{Center for Human-Compatible AI, UC Berkeley, Berkeley, USA}

\icmlaffiliation{rome}{Institute of Cognitive Sciences and Technologies, National Research Council, Rome, Italy}

\icmlcorrespondingauthor{Axel Abels}{axel.abels@ulb.be}
\icmlkeywords{Machine Learning, ICML, Collective Intelligence, Collective Decision-Making, Bias, Expertise Trees, Expert}

\vskip 0.3in
]

\printAffiliationsAndNotice{}

\begin{abstract}
Experts advising decision-makers are likely to display expertise which varies as a function of the problem instance. In practice, this may lead to sub-optimal or discriminatory decisions against minority cases. In this work we model such changes in depth and breadth of knowledge as a partitioning of the problem space into regions of differing expertise. We provide here new algorithms that explicitly consider and adapt to the relationship between problem instances and experts' knowledge. We first propose and highlight the drawbacks of a naive approach based on nearest neighbor queries. To address these drawbacks we then introduce a novel algorithm --- expertise trees --- that constructs decision trees enabling the learner to select appropriate models. We provide theoretical insights and empirically validate the improved performance of our novel approach on a range of problems for which existing methods proved to be inadequate. 

\end{abstract}

\section{Introduction}
The following example from medical diagnostics illustrates the problem settings we examine in this work: an online platform repeatedly presents a group of medical experts with patients demonstrating sets of symptoms. Based on these symptoms, the patient's medical record, and the experts' knowledge and experience, the most appropriate treatment must be selected. This treatment is then applied to the patient, and its effectiveness is evaluated based on some measure of recovery, side effects, and cost, for example. A rational objective in this setting is to maximize the effectiveness of the treatments administered in function of the advice of the group of medical experts. 
Ideally, we would hope that medical experts agree on the best treatment based on objective characteristics of each treatment, such as its success rate, side effects, and cost.
However, expert opinions are likely to be influenced by both objective factors and subjective views.
For example, based on previous experience with comparable patients, a clinician may have views about the efficacy of a treatment for a patient that differs from another clinician's views. Such divergences can result from factors which are clearly outside of an expert's control, but they can also be the result of cognitive biases. 
For example, medical experts might overstate the efficacy of a treatment as a result of confirmation bias \cite{trope1997wishful} or the primacy effect \cite{bond2007information}.
 Experts thus provide individual advice based on their previous experience with similar problems. When we have access to a diverse set of experts who might disagree, we wish to determine on which advice to act. Particularly, we consider the case wherein multiple iterations (on different problem instances, e.g., different patients with different medical records) of the decision-making process occur, allowing us to adapt to expertise. The challenge of identifying an appropriate way to act on expert advice is heightened by the partial feedback: we only observe the outcome of the decisions we make, e.g., the chosen treatment.

The use of expert advice has been extensively studied to solve complex decision problems in the face of uncertainty, known as \emph{bandits} \cite{auer2002finite,foster2020beyond}. Prior approaches to leveraging expert advice in this setting are however not conditioned on the characteristics of each problem instance. In contrast, we consider here \emph{localized expertise}, specifically the case wherein at each iteration the quality of experts may change in function of the problem instance, i.e., an observed set of features. Such varying levels of expertise are normal for human experts who specialize in particular areas of a larger problem domain. Similarly, artificial experts may exhibit performance fluctuations as a consequence of insufficient training, or of incomplete or biased data sets.
Specifically, (un)conscious biases may result in judgmental errors caused by an erroneous weighting of sensitive traits such as gender or ethnicity \cite{pohl2017cognitive}. 
Such biases can occur both in humans and in artificial experts as a result of biased design or data bias \cite{gianfrancesco2018potential}.
 Methods capable of detecting and counteracting such biases are not only essential in terms of performance, but also in terms of fairness.
 Concretely, we explore the improvements in performance that can be achieved when accounting for changes in expertise in function of a set of features, for example as a result of specializations or biases. 
Thus, while there may not be a globally optimal way of acting on expert advice, partitioning the problem space and developing partial models on these partitions may provide an effective piece-wise model. In this paper, we argue that algorithms designed for collective decision-making must account for localized expertise to maximize performance. In particular, algorithms for bandits with expert advice should learn policies conditioned on the problem instance. 
 
 To address this need, we introduce two novel algorithms, based first on a nearest neighbors approach, and second on a model tree approach. To highlight the difference with model trees we name the latter approach \emph{expertise trees}. The increasing complexity of these approaches is required to tackle the challenges of localized expertise. We then provide some bounds on the performance of algorithms in the localized setting and provide theoretical insights into the losses incurred by inappropriately partitioning the problem space. To conclude, we empirically evaluate these algorithms in terms of performance and execution time and show that the expertise tree approach achieves the best results, at an acceptable increase in algorithmic complexity.

\section{Bandits with Localized Expert Advice} 

\begin{figure*}
\centering
 
 \includegraphics[width=.7\textwidth]{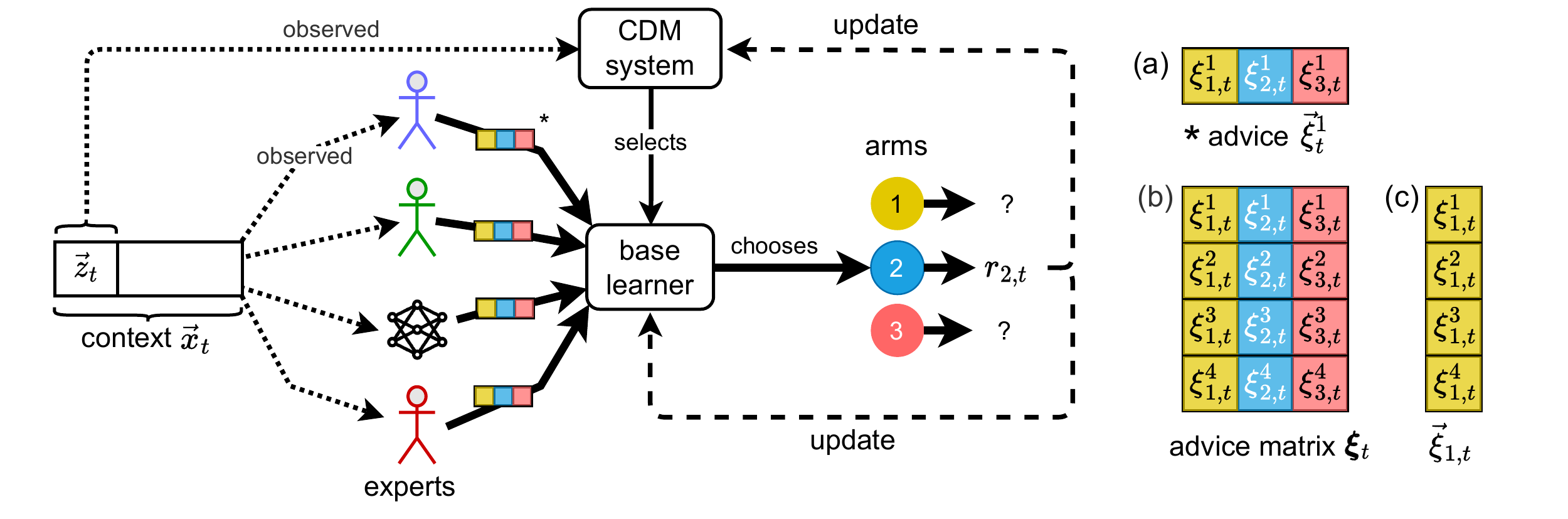}

 \caption{%
  Illustration of a bandit with localized expertise with $N=4$ experts and $K=3$ arms.
  The CDM system observes the expertise context $\vec{z}_t$ and selects an appropriate localized learner. 
  Each expert observes the full context $\vec{x}_t$ at time $t$ and then provides an advice vector to the learner (e.g., (a) $\vec{\xi}^1_t$ is the first expert's advice). Note that one can have both human and artificial experts as depicted in the figure. The advice matrix  (b) $\boldsymbol{\xi}_t$ is a concatenation of these advice vectors, and  (c) $\vec{\xi}_{1,t}$ is the advice for arm $1$.  The chosen arm ($k=2$) provides a reward $r_{2,t}$ sampled from a distribution with mean $f(2,\vec{x}_t)$. This reward is then used to update both how the learner acts on advice, and how the CDM system chooses an appropriate learner. }\label{fig:problem-illustration}
\end{figure*}

Bandits formalize problems wherein a learner repeatedly chooses one out of $K$ arms over a number of rounds with the aim of optimizing the outcomes of its choices when only the outcome of the chosen arm is observed \cite{ThompsonONTL,auer2002finite}. In the contextual setting \cite{chu2011contextual,DBLP:journals/corr/ValkoKMFC13}, outcomes are characterized by a function $f: [K] \times \mathds{R}^d \rightarrow [0,1]$, which maps an arm $k$ and the context $\vx_t$ (i.e., a set of features characterizing the current instance) of the decision at time $t$ to an expected reward. At time $t$, the reward observed for choosing arm $k_t$ in function of a context $\vx_t$ is $r_t = f(k_t,\vx_t) + \epsilon$, with $\epsilon$ a noise term with mean $0$. For example, the treatment (arm) with the best outcome (reward) in terms of some relevant measure, such as QALY \cite{whitehead2010health}, varies in function of a patient's medical records (the context).
When the complexity of the problem is such that the cost of learning is unacceptable, a learner can exploit the advice of experts to make decisions, as formalized by the problem of \emph{bandits with expert advice} \cite{auer2002finite,foster2020beyond}.
In particular, when approximating $f$ from scratch is unfeasible, the learner can observe the advice of a set of $N$ experts in the form of $N$ estimates of each arm's expected reward. We denote by $\vv{\xi}^n : \mathds{R}^d \rightarrow \mathds{R}^K$ expert $n$'s vector function of $K$ estimates, one per arm. 

To maximize the rewards from the arms it chooses, a learner must optimize its decisions in function of these advice vectors. Algorithms for bandits with expert advice therefore learn to act on the advice matrix by maintaining a policy $\pi_t$ parameterized by the matrix of advice $\boldsymbol{\xi}(\vx_t) := \{\vv{\xi}^1(\vx_t),...,\vv{\xi}^N(\vx_t) \}$.
For the sake of readability, we drop the parameters of $\xi$ when no ambiguity exists, e.g., we will write $\boldsymbol{\xi}_t$ as shorthand for the $N\times K$ matrix of advice induced by the context at time $t$. Similarly, $\vv{\xi}_{k,t}$ denotes the $k$-th column of this matrix, i.e., the advice for arm $k$.
In every round $t$, the policy $\pi_t$ induces a choice of arm $k_{\pi_t}$ for which a reward $r_{\pi_t}$ is observed. The aim of the learner is to update its policy in function of the observed rewards to maximize the average of collected rewards, $\frac{1}{T}\sum_{t=1}^T r_t$. Given an optimal policy $\pi^*$, performance can also be stated as the expected gap to this policy, i.e., the regret:  $R(T) = \mathds{E} [\frac{1}{T}\sum_{t=1}^T r_{\pi^*} -  r_{\pi_t}]$, where the expectation is over possible contexts and resulting outcomes.

\subsection{Localized Expertise}

We generalize the problem by explicitly considering cases wherein the quality of experts' advice changes in function of the context they are presented with.
This is naturally present for human experts who tend to have an area of specialization. When the instance falls outside of an expert's specialization, this should be accounted for. Similarly, artificial experts may be well-fitted to only a subset of the problem space. Biases, whether conscious or not, might also cause inaccuracies induced by some sensitive features, such as gender or ethnicity. Identifying the dependence of expert quality on a set of features has the potential to significantly increase the quality of the collective decisions. We hypothesize that we can obtain an effective model by partitioning the context space and learning partial models on these partitions. 

We formalize this dependence by introducing the \emph{expertise context}, which is defined as a subset of the full context observed by the experts $\vv{x}_t$. For example, a patient's gender might affect the effectiveness of treatments, but also (unconsciously) affect the quality of expertise. While we might reasonably want to include the full context $\vv{x}_t$ when deciding on which expertise to act, this subset formulation accounts for cases wherein some features cannot be meaningfully captured. For example, textual descriptions of the problem can readily be incorporated by human experts, but the computational cost might be too high for them to be taken into account by an algorithm for bandits with expert advice. We denote the subset of dimension $g\leq d$ by $\vv{z}_t\in \mathds{R}^g$. In the localized expertise setting, the optimal way of acting on expert advice is thus conditioned on this subset of features. For example, if a patient's symptoms relate to the heart, the advice of medical experts with matching specialization should be prioritized. In practical terms, $\vv{z}_t$ encodes a patient's category, and the decision-maker should optimize the use of expert advice in function of the category encoded in $\vv{z}_t$. In particular we are interested in settings wherein the relation between the expertise context $\vv{z}$ and experts is unknown a priori, and must thus be learned through feedback from decisions.

\subsection{Reduction to a Contextual Bandit}\label{sec:cmetamab}
As a baseline, we consider selecting a single expert whose advice we act on at each timestep. This requires determining the expert whose advice leads to a maximized expected reward given the observable expertise context $\vv{z}_t$. 
In practice, we are reducing the problem to a contextual multi-armed bandit. Specifically, each arm in the constructed bandit corresponds to an expert in the original problem, and selecting that arm implies following the corresponding expert's advice. The reward we observe for the chosen arm allows us to then update the quality estimate of the expert whose advice we acted on.
The quality of each expert (i.e., arm in the constructed bandit) in turn is a function of the expertise context. We can learn to approximate this function while ensuring sufficient exploration by applying an appropriate contextual bandit algorithm to the reduced problem. 
 This approach can be seen as a generalization of a meta multi-armed bandit approach (\cite{auer2002finite}, which uses a standard multi-armed bandit algorithm to determine the best expert by modelling each expert as a an arm in a meta-bandit) to the contextual case. 
 
 The chosen contextual bandit algorithm impacts the specializations that can be learned. For example, algorithms based on decision trees (e.g., \emph{TreeHeuristic} \cite{DBLP:journals/corr/ElmachtoubMOP17}) are more suitable for discrete expertise regions. An additional benefit of tree-based algorithms is that they perform feature selection, which is essential if the expertise context contains many features of which only a few are predictive of expertise.

One significant drawback of this reduction is that decisions are taken on the basis of a single expert's advice. In addition to limiting knowledge acquisition in each round to a single expert --- which typically induces poor performance when the number of experts is high --- it is also impossible for the learner to surpass the performance of the best individual in the collective for any given expertise context. In contrast, when a decision is taken as a function of the advice of multiple experts, an opportunity for collective intelligence (i.e., performance better than the single best expert) emerges. With this goal in mind, we propose in the following section  (i) a nearest neighbor approach inspired by local learning and highlight its drawbacks, and then (ii) address these drawbacks with our novel expertise tree approach. 

\section{Methods}\label{sec:methods}

Our goal is to learn a mapping from expert advice to arms in such a way that our decisions maximize the cumulative reward. Because expert quality is a function of the expertise context $\vv{z}_t$, our aim is to learn a mapping conditioned on $\vv{z}_t$, i.e., we hope to learn a policy $\mathcal{\pi}: [0,1]^{ N} \times \mathds{R}^{g \times K} \rightarrow \Delta^K$ which maps both the advice and the expertise context to a distribution over the arms.
We propose to learn $\pi$ by partitioning the expertise space into regions, and maintaining for each region an independent instance of a bandits-with-expert-advice learner, such as EXP4 \cite{auer2002finite} or Meta-CMAB \cite{abels2023dealing}.

If prior knowledge is available on which features of $\vv{z}_t$ are likely to alter expert performance, this can be used to partition the space appropriately and then apply an independent algorithm for bandits with expert advice on each subset. If no such knowledge is available, or if the influence of specific features is over or underestimated, the chosen partitions will reflect these inaccuracies. When no knowledge about appropriate partitions is available a priori, a sensible solution is to act in function of previous similar experiences.

We thus consider an algorithm which adapts the principles of local learning  \cite{bottou1992local} to the bandits with expert advice setting.
 When given an expertise context $\vv{z}_t$ at time $t$, we can sample a percentage $p$ of previous experiences for which the expertise context is closest. These $\lceil \frac{p}{100}\cdot t \rceil$ experiences are then used to initialize a bandits-with-expert-advice learner. We refer to this approach as \emph{Nearest $p\%$}. While this method requires no knowledge of the expected number of partitions, a proper choice of $p$ should be made to avoid over or under-fitting. Alternatively, a distance threshold can be used instead of a threshold on the number of neighbors. In both cases, too large thresholds will include previous experience from different regions, while too small values can induce an under-fit model. Perhaps as significant as the choice of threshold is that the performance of this approach relies on the accuracy of the distance measure as a proxy for model similarity. In particular, if the expertise context contains features which have no bearing on expert quality, their inclusion in the distance computation will impact the quality of the learned model. To cope with such uncorrelated features, and to automatically learn appropriate neighborhoods, we propose in the following section a novel approach which simultaneously learns to identify relevant features and partitions through a decision tree.

 \subsection{Expertise Trees to Capture Localized Knowledge}\label{sec:expertise_tree}
 
Model Trees \cite{wang1996induction} are decision trees wherein a regression model (typically linear) is fit on the leaves. This kind of decision tree effectively partitions the space into regions for which an individual model is beneficial. We develop here a model tree algorithm adapted to the setting of bandits with localized expert advice. This setting differs from the usual model tree setting in two aspects. First, the bandit setting induces a need for exploration and a different optimization target (reward as opposed to prediction error). Secondly, for traditional model trees, the features on which the space is split are the same features that are used in the leaf models. In contrast with model trees, we split the expertise space (i.e., on $\vv{z}_t$) and learn models on the expert advice (i.e., $\boldsymbol{\xi}_t$). Given this difference, we label these \emph{expertise trees}.

\subsubsection{Building Expertise Trees}\label{sec:building_tree}
Let $\mathcal{H}$ be a history of experience tuples consisting of the advice matrix, the chosen action, the collected reward, the probability for that action, and the expertise context, i.e., a $\langle\boldsymbol{\xi}, k, r,p,\vv{z}\rangle$ tuple.
Given $\mathcal{H}$, we can partition this set of experiences along the $i^{th}$ dimension of the expertise contexts, resulting in two subsets we will denote as $\mathcal{H}_{<\tau^{(i)}} := \{\langle\boldsymbol{\xi}, k, r,p,\vv{z}\rangle \in \mathcal{H} | z^{(i)}<\tau^{(i)}\}$, and analogously $\mathcal{H}_{\geq\tau^{(i)}}:= \mathcal{H}\setminus \mathcal{H}_{<\tau^{(i)}}$, wherein $z^{(i)}$ denotes the $i$th value of the expertise context $\vv{z}$ and $\tau^{(i)}$ is the threshold at which we partition. These two experience subsets can then be used to maintain two algorithms for bandits with expert advice, which we will denote respectively by $\mathrm{ALG}_{<\tau^{(i)}}$ and $\mathrm{ALG}_{\geq\tau^{(i)}}$. Let $Q(\mathrm{ALG},\mathcal{H})$ be an estimate of the performance of some algorithm $\mathrm{ALG}$ on the set of experiences $\mathcal{H}$, to be defined more precisely in the following section. We only benefit from a split if the two subset learners perform better than the single learner $\mathrm{ALG}$ trained on $\mathcal{H}$, which is the case if 
\begin{multline}
    Q(\mathrm{ALG}_{<\tau^{(i)}},\mathcal{H}_{<\tau^{(i)}})|\mathcal{H}_{<\tau^{(i)}}| + \\ Q(\mathrm{ALG}_{\geq\tau^{(i)}},\mathcal{H}_{\geq\tau^{(i)}})|\mathcal{H}_{\geq\tau^{(i)}}| > 
     Q(\mathrm{ALG},\mathcal{H})|\mathcal{H}|
     \label{eq:criteria_split}
\end{multline}
Note that the two quality estimates are weighted by the sizes of their respective subsets.

 \begin{algorithm}[tb]
 \begin{algorithmic}
 \REQUIRE $N$ experts, contextual bandit with reward function $f: [K] \times \mathds{R} \rightarrow [0,1]$ 
 \STATE  $root \gets$ Initialize empty expertise tree
 \FOR{$t = 1, 2, ..., T$}
 \STATE Experts observe the context ${\vv x}_t$ 
 \STATE Get the observable subset of ${\vv x}_t$, i.e., the expertise context ${\vv z}_t$ 
 \STATE {Get expert advice $\pmb{ \xi}_t=\{\vv{{\xi}}^1({\vv x}_t), ..., \vv{\xi}^N({\vv x}_t)\}$ } 

 \STATE $node \gets root$
 \WHILE {$node$ has a beneficial split}
    \STATE $node \gets $ child of $node$ containing $\vv{z}_t$
 \ENDWHILE
 \STATE Let $\pi_t$ be the policy of $node$'s learner
 \STATE Pull arm $k_{t} \sim {\pi}_t(\pmb{ \xi}_t)$ and collect resulting reward $r_t$
 \STATE $p \gets {\pi}_{k_t,t}(\pmb{ \xi}_t)$ \COMMENT {get probability of chosen arm} 
    \IF {not incremental}
        \STATE $node \gets root$ \COMMENT{update tree starting from its root}
    \ENDIF
    \STATE update $node$ with $\langle {\pmb{ \xi}_t}, k_t,r_t,p\rangle$
    \WHILE {node is not empty}
    \STATE update candidate splits rooted in $node$ with $\langle {\pmb{ \xi}_t}, k_t,r_t,p\rangle$
    \IF {the best split candidate is such that Equation (1) holds}
        \STATE split $node$
    \ENDIF 
    \STATE set $node$ to child containing $\vv{z}_t$
        
    \ENDWHILE
 \ENDFOR
 \end{algorithmic}
 \caption{(Incremental) ExpertiseTree}\label{alg:expertisetree}
 \end{algorithm}

\subsubsection{Estimating Expertise Quality}\label{sec:model_quality}

 Given a set of experiences $\mathcal{H}$ it is useful to ask the question how another policy would have performed in terms of average reward.
 Following \citeay{doi:10.1137/S0097539701398375}, we can estimate this expected average reward of a learner $\mathrm{ALG}$ and its policy $\pi^{\mathrm{ALG}}$ on that set of experiences as 
\begin{equation} 
    Q(\mathrm{ALG},\mathcal{H}) =  \frac{1}{|\mathcal{H}|}\sum_{\langle\boldsymbol{\xi}, k, r,p,\vv{z}\rangle \in \mathcal{H}} \pi^{\mathrm{ALG}}_{k}(\boldsymbol{\xi}) r/p \label{eq:model_quality}
\end{equation}
Wherein $\pi^{\mathrm{ALG}}_{k}(\boldsymbol{\xi})$ denotes the policy's probability for action $k$ given the advice vector $\boldsymbol{\xi}$.
Because these experiences are acquired by a policy (and thus biased), the probability in the denominator ensures the expected reward is unbiased. 

We consider therefore the following approach: build an expertise tree wherein the features on which we split are the features of $\vv{z}$, then run an algorithm for bandits with expert advice on the experiences in the leaves. 
More specifically, while a beneficial split exists (i.e., a split for which (\ref{eq:criteria_split}) holds), we repeatedly split the expertise space into subsets and maintain learners specific to those partitions. Among splits for which (\ref{eq:criteria_split}) holds, we select the split which maximizes the left-hand side of the inequality. Pseudocode is provided in Algorithm  \ref{alg:expertisetree}. This approach effectively combines algorithms for bandits with expert advice with the flexibility of decision trees. The result is a two-step model, on one hand, the decision tree on $\vv{z}$, and on the other hand the model in each leaf. This effectively splits the space of $\vv{z}$ into regions, each with its own policy.

 As each potential split requires simulating the run of two algorithms for bandits with expert advice, considering all possible splits for continuous variables becomes intractable. Instead, following \citeay{potts2005incremental}, we fix $\kappa$ split candidates for each feature. While this slightly reduces flexibility, it also greatly reduces computation time. 

\subsubsection{Incremental Expertise Trees}
Because retraining an expertise tree for each experience is expensive, we propose an  algorithm which constructs the tree incrementally, i.e., at each time step, models are updated with regards to the collected experience, and splits are made if they improve the quality of the expertise tree. In contrast with the non-incremental variants, splits, once made, are not revisited. As a result, if the stochasticity of the problem induces a bad split, it can impact performance going forward.
Note that the variance of the quality estimate  \eqref{eq:model_quality} is such that heuristics which attempt to limit early splits with the use of for example Hoeffding bounds \cite{pfahringer2007new} are of limited use. However, the additional exploration (as discussed in Section \ref{sec:theory}) induced by a split acts as an inhibitor or unnecessary splits. 

The trade-off for a possible performance decrease is a significantly more efficient algorithm. Fully learning an expertise tree is more time-intensive than incrementally updating an expertise tree. More specifically, assuming an ideal tree depth of $h$ (which is, in turn, logarithmic in the number of partitions), the dimension of the expertise context $g$, and the number of candidate splits per feature $\kappa$, building the corresponding expertise tree requires estimating $O(2^h g \kappa)$ models. Updating the expertise tree requires $O(hg\kappa)$ updates ($g\kappa$ updates for each node on the path to the leaf in which the expertise context resides). In contrast, updating an incremental tree requires only $O(g\kappa)$ model updates on average, as only leaf models are updated.

\section{Theoretical Results}
In the following, we provide some bounds on the problem of localized expertise, and establish the cost of inappropriately partitioning the context space.

Let $\mathcal{Z}$ be regions of expertise, i.e., a partition of $\mathds{R}^g$ wherein expertise is homogeneous within each subset, in other words there is a single optimal policy for each subset.

\textbf{Cost of Localized Expertise} We first establish a lower bound on the regret of algorithms for bandits with localized expertise. Assume $\mathcal{Z}$ is known, such that we can run $|\mathcal{Z}|$ instances of a bandit with expert advice algorithm and assign each experience with expertise context $\vec{z}$ to the corresponding learner. Denote by $p(Z)$ the probability that a sampled context is contained by the region $Z\in\mathcal{Z}$. Let $R(T)$ be the regret incurred by an instance of the bandit with expert advice algorithm. The total regret incurred by all instances is then $\sum_{Z \in \mathcal{Z}}R(p(Z)T)$. In the usual case wherein $R(T) \propto \sqrt{T}$ this measure is maximized when the entropy of $p$ is maximized; $p(Z) = 1/|\mathcal{Z}|\; \forall Z \in \mathcal{Z}$, leading to a regret of $ |\mathcal{Z}| R(T/|\mathcal{Z}|)$. And conversely, it is minimized when $\max_{Z \in \mathcal{Z}} p(Z) = 1$; which is equivalent to having no localized expertise, thus reducing the regret to $R(T)$.

\textbf{Expected Regret of Non-Localized Algorithms}
For any expert $n$, let $p'(n)$ be the proportion of expertise contexts for which expert $n$ is optimal. It then follows that any non-specialized algorithm, i.e., any algorithm that selects the same expert regardless of context, has a probability of at most $p'(n^*) = \max_{n \in [N]} p'(n)$ of selecting the best expert in each region. In particular, the algorithm will incur $0$ regret for contexts for which the expert is optimal, and constant regret for all other contexts. The resulting lower bound on the regret is thus $\Omega((1-p'(n^*))T)$. Any non-specialized algorithm must therefore incur linear regret in the localized setting. This lower bound is maximized when the entropy of $p'$ is maximized, in which case the regret incurred is $\Omega(\frac{N-1}{N}T)$.

\textbf{The cost of (not) splitting}\label{sec:theory}
Let $\mathcal{H}$ be a set of experiences to be split, and $\tau^{(i)}$ the threshold on the $i$-th feature. Let $\mathcal{Q} = |\mathcal{H}|Q(ALG, \mathcal{H})$ be shorthand for the expected cumulative reward of a learner on the parent with history $\mathcal{H}$ as defined in \autoref{eq:model_quality}. Similarly, we let the performance on the left and right child be respectively 
$\mathcal{Q}_l = |\mathcal{H}_{< \tau^{(i)}}|Q(ALG, \mathcal{H}_{< \tau^{(i)}})$ and $\mathcal{Q}_r = |\mathcal{H}_{\geq \tau^{(i)}}| Q(ALG, \mathcal{H}_{\geq \tau^{(i)}})$.

A split is necessary if $\mathcal{Q} < \mathcal{Q}_l + \mathcal{Q}_r$. In other words, if the performance is improved by maintaining two policies.

Let $\mathcal{p}$ be the probability that a context is contained by the left child if it is contained by the parent. Conversely, the probability that an experience is contained by the right child is $1-\mathcal{p}$.
 We then have that the expected regret of the split is $R(\mathcal{p}T)+R((1-\mathcal{p})T)$. 
Where $\mathcal{p}T$ and $(1-\mathcal{p})T$ are respectively the expected number of experiences assigned to the left and right child.

The performance on the unsplit node can be expressed as a function of the gap between the expected cumulative reward on the unsplit node and the weighted sum of expected cumulative rewards on the split.

In particular, let $\mathcal{R}=  \max_\pi \mathds{E}[ f(k_{\pi(\boldsymbol{\xi}(\vx))},\vx)]$, where the expectation is over the context sampled from the relevant region of the context space, be the expected reward of the best policy on experiences contained by the parent node. And let $\mathcal{R}_l$ and  $\mathcal{R}_r$ analogously be respectively the expected maximum reward in the left and right child. 

We then let $\Delta =  \mathcal{p}\mathcal{R}_l + (1-\mathcal{p})\mathcal{R}_r - \mathcal{R} $  be the gap in expected average reward between the unsplit or split learners. 
By the definition of regret, we have with expectation that the cumulative reward of a learner on the parent node's experiences is $\mathds{E}[\mathcal{Q}] = \mathcal{R}T - R(T) $, i.e., the optimal performance minus the regret.
Similarly, the expected performance on the children's experiences is $\mathds{E}[\mathcal{Q}_l] =  \mathcal{R_l}T\mathcal{p} - R(T\mathcal{p}) $, and $\mathds{E}[\mathcal{Q}_r] =  \mathcal{R_r}T\mathcal{(1-p)} - R(T(1-\mathcal{p})) $

We can then formulate the expected improvement of a split:
\begin{align*}   
&\mathds{E}[\mathcal{Q}_l + \mathcal{Q}_r - \mathcal{Q}] \\ &= (\mathcal{R_l}\mathcal{p} + \mathcal{R_r}\mathcal{(1-p)} - \mathcal{R})T  -    R(T\mathcal{p}) - R(T(1-\mathcal{p})) \\ &\quad + R(T)  \\
 &= \Delta T - (   R(T\mathcal{p}) + R(T(1-\mathcal{p})) - R(T) )  \\&= \Delta T -  (\sqrt{\mathcal{p}}+\sqrt{1-\mathcal{p}}-1)R(T)   \end{align*}
 Where the last equality assumes that $R(T) \propto \sqrt{T}$ (as is typical for expert advice algorithms, see for example EXP4-IX \cite{neu2015explore} or Meta-CMAB \cite{abels2023dealing}).

In particular this means that a split is only beneficial if the gap in performance is sufficiently large. This is a consequence of the additional exploration induced by the split. Thus, if experts are sufficiently close, i.e. if $\Delta <  (\sqrt{\mathcal{p}}+\sqrt{1-\mathcal{p}}-1)R(T)/T $ a split is detrimental. 
As $R(T)/T$ typically is monotonically increasing in both the number of arms $K$ and the number of experts $N$, but is monotonically decreasing in the number of timesteps $T$, this confirms the intuition that splits are most beneficial when $T$ is large or when $K$ and $N$ (and the exploration cost they induce) are small.

A special case of this is when $\Delta=0$, i.e., when the optimal policy in the unsplit node is identical to the policy in the child nodes. In such a case all learners converge towards the same policy, but the child nodes perform some additional exploration, resulting in degraded performance. In particular, this additional exploration penalty is $R(T\mathcal{p}) + R(T(1-\mathcal{p})) - R(T)$.
 As reasonable algorithms for bandits with expert advice display a sub-linear dependence on $T$, the resulting sum is positive.  And thus, the split incurs a regret proportional to $\sqrt{pT} + \sqrt{(1-p)T}= (\sqrt{p}+\sqrt{1-p})\sqrt{T}$. Which is $(\sqrt{p}+\sqrt{1-p})$ times larger than the regret of the parent node. In particular, this sum is largest when $p=0.5$, in which case the unnecessary split magnifies the regret by $\sqrt{2}$.

\section{Experimental Setting}
We evaluate our algorithms in terms of average reward. Throughout we compare the performance of the following algorithms: \emph{Meta-CMAB}, a state-of-the-art approach for bandits with (non-localized) expert advice; \emph{TreeHeuristic} \cite{DBLP:journals/corr/ElmachtoubMOP17} applied to our reduction presented in Section \ref{sec:cmetamab}; \emph{Oracle}, which relies on prior knowledge of the optimal partition and maintains one independent instance of Meta-CMAB per region; \emph{Nearest $1\%$} and $10\%$, which take decisions based on models trained with the nearest $1\%$ or $10\%$ of past experiences; Finally, the \emph{(Incremental) Expertise Tree} algorithm, as presented in Section \ref{sec:expertise_tree}. 
Our results are averaged over $100$ simulations, and for a varying number of arms and experts. The code to reproduce these results is provided in the supplementary material.

\subsection{Bandit Problems}

Following \citeay{riquelme2018deep}, we build bandits out of classification problems. As a result, for a given context, a single arm (corresponding to the true label) delivers a reward of $1$, and the other arms provide $0$ reward.
We evaluate on a variety of datasets presenting a diversity of feature distributions, arm counts, and reward distributions over the arms, chosen from the openml data repository \cite{vanschoren2014openml}. The dataset selection and processing is provided in the supplementary information.

\subsection{Simulating Localized Expertise}

\begin{figure}
\centering
     \includegraphics[height=.17\textwidth]{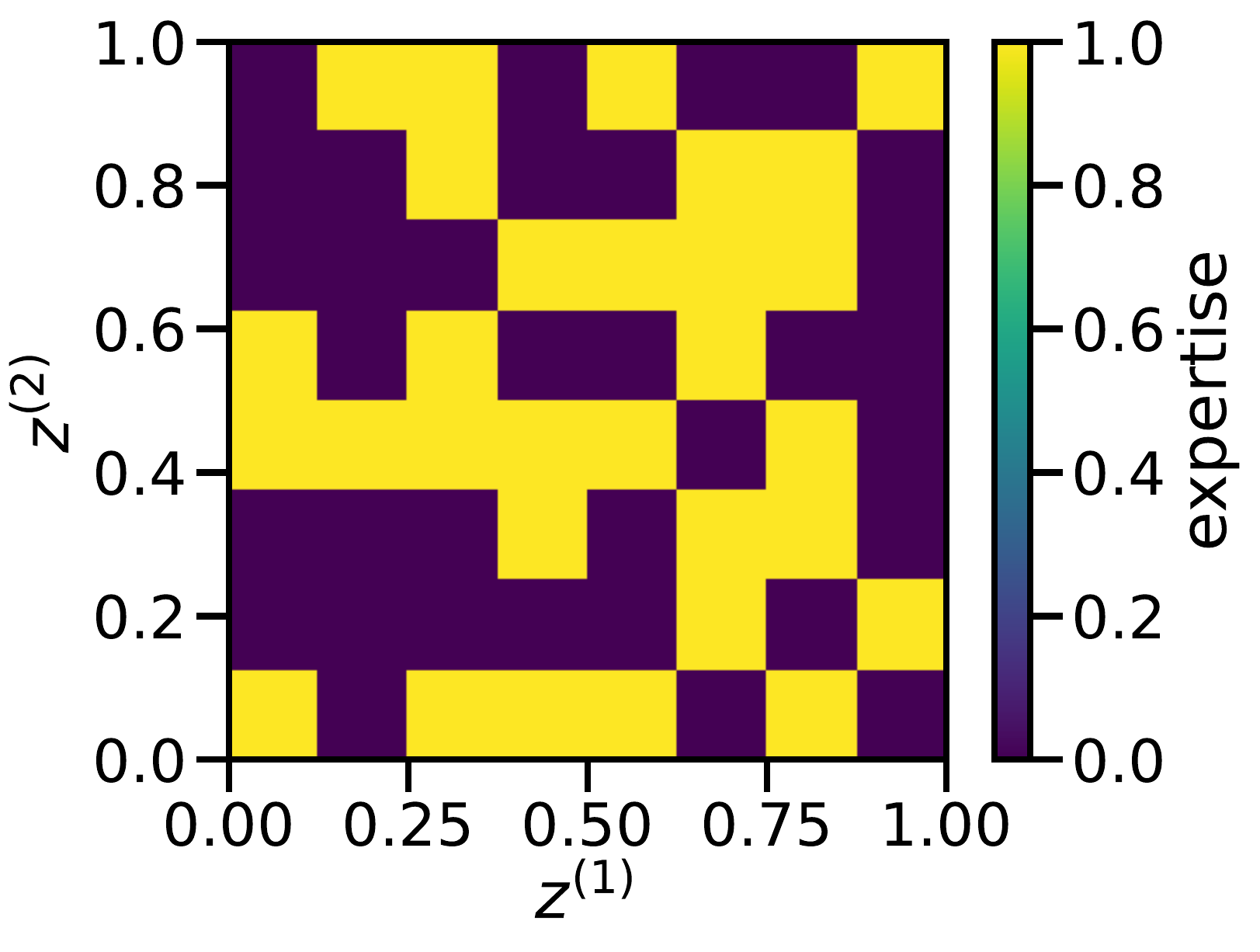}

     \caption{%
      Example of an expertise heatmap with $64$ regions. Heatmaps are generated by randomly assigning an expertise value of $0$ or $1$ to each cell. Each time step one point in this space is sampled and the expertise of each expert is determined by their individual heatmap. For high expertise values, experts provide honest advice, for low values they provide adversarial advice.  %
   }\label{fig:heatmaps}
\end{figure}

We consider changes in expertise characterized by expertise heatmaps which map expertise contexts to expert quality. Each round, experts advice on a context vector $\vx_t$ from which a subset of $g$ features (chosen randomly at the start of the experiment) form the expertise context $\vv{z}_t$. Among these $g$ features, $2$ features are correlated with expertise, as illustrated in Figure $\ref{fig:heatmaps}$. An appropriate learner thus has to learn both which of the $g$ features is relevant, and simultaneously learn an appropriate partition. Each expert is assigned an individual heatmap, and for a given value of $\vv{z}_t$, the expert's quality is determined by the corresponding expertise value. Values close to $0$ result in adversarial advice, while high expertise results in honest advice. Heatmaps are obtained by assigning for each expert an expertise value of $0$ or $1$ to each of the regions (either $1$, $4$, $16$ or $64$ regions). Increased number of regions increases the difficulty of the problem. For example, to optimize performance in the hardest configuration we consider, $64$ regions must be correctly identified based on a limited number of experiences.
By increasing the size of $g$, we make the problem more complex by increasing the number of uncorrelated features. 

\section{Results and Discussion}

We show results in terms of average reward instead of regret as this allows us to display both the loss incurred by increasing the number of regions, as well as the loss in performance of individual algorithms. When presenting results in terms of regret, the former information would be obscured as this loss is incurred by both the oracle and the individual algorithms. Results are averaged over $T=1000$ steps for all expert counts ($N \in \{4,32\}$) and dataset combinations.

\subsection{Expertise Tree Algorithms are More Robust to Increases in Problem Complexity}
Learning in this setting is challenging in two ways, as both the appropriate regions must be identified, and for each region, an appropriate model needs to be learned. To isolate the loss of performance resulting from the need to identify these regions, we compare the performance of our algorithms to an Oracle algorithm. For this algorithm, we assume that perfect knowledge of the regions is known, hence any loss in performance of the Oracle variant results only from uncertainty about the appropriate model for each region. This allows us to isolate how much performance loss of non-oracular variants results from uncertainty about the structure of the expertise space.

\begin{figure*}
\centering
     \includegraphics[width=1\textwidth]{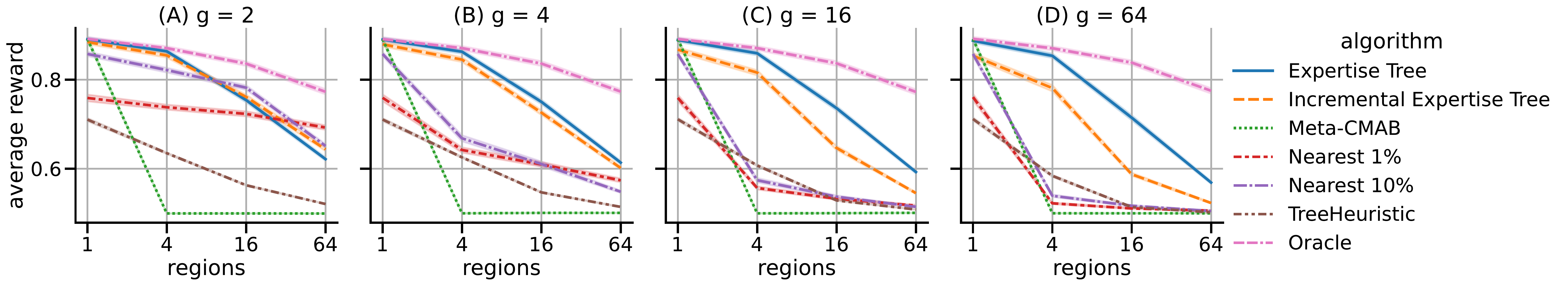}

     \caption{%
      Average reward in function of the number of regions (x-axis) and expertise context size ($g$, columns) for different algorithms. Shaded areas show the $95\%$ confidence interval.
   }\label{fig:performance_region}
\end{figure*}

Figure \ref{fig:performance_region} illustrates the change in performance of our algorithms as the number of regions increases. These plots show that when specializations are present (i.e., more than $1$ region), the non-specialized Meta-CMAB algorithm fails to learn. These plots also show that, as the number of regions increases, the performance of specialized algorithms declines more rapidly than the performance of Oracle. This is a consequence of the uncertainty about the partitioning of the expertise context space. This is because appropriate partitions for a high number of regions require several optimal splits. Figure \ref{fig:performance_region} also shows that, for a given number of regions, the size of the expertise context makes it harder to identify the ideal partition, again illustrated by the growing gap with the performance of the Oracle. This is because random correlations can occur when the number of features is larger, inducing splits without actual discriminating power. Although this causes all specialized algorithms to degrade, it is less pronounced for both expertise tree variants.
The incremental variant of the ExpertiseTree suffers more from the uncorrelated features as splits resulting from random correlations cannot be reversed. 

Nearest Neighbor models trained on either $1$ or $10\%$ of experiences assume that the expertise space is divided into approximately $100$ or $10$ regions. As the number of regions drifts towards the former, the performance of the $1\%$ variant improves relative to the $10\%$. In addition, as the size of the expertise context increases, the quality of the distance metric as a proxy for similarity in terms of expert quality decreases. This results in a decline in the performance of the Nearest Neighbor algorithms, even when the number of regions is small.

Finally, while the TreeHeuristic algorithm can perform the same feature selection and partitioning as the ExpertiseTree algorithm, it is penalized by its single expert approach. This not only impacts the best decision that can be made by TreeHeuristic at any timestep, but it also impacts its updates, as each experience is used to update a single expert's quality estimate. This effectively means that TreeHeuristic has $N$ fewer experiences per expert than our methods.

Note that, for all algorithms, including the Oracle, a loss of performance relative to a non-localized setting is inevitable. More regions require more individual models, but, given the constant number of experiences, each individual model is trained less, leading to poorer performance.

\subsection{Localized Algorithms on Global Expertise}
If expertise is not localized, localized models are likely to perform worse, as they take fewer relevant experiences into account.
The leftmost point of each sub-figure in Figure \ref{fig:performance_region} illustrates the performance of various algorithms when experts are not localized (i.e., when there is a single region). Notably, the performance of TreeHeuristic is significantly inferior to other algorithms. This performance loss results from TreeHeuristic's policy which selects one expert per learned partition. As a result, its performance is bounded by the performance of the best expert. This is a limitation not shared by the other algorithms. It is however noteworthy that the Nearest Neighbor algorithms show some loss in performance compared to the non-specialized Meta-CMAB. Because these algorithms learn a model on a subset of all experiences, the learned model generalizes less well, leading to a loss of performance. This is particularly the case when only $1\%$ of experiences are taken into account. In contrast, the ExpertiseTree approach will maintain a single partition, as no beneficial splits exist. The performance of the ExpertiseTree is almost identical to the simple Meta-CMAB, even when the size of the expertise context increases significantly. The Incremental ExpertiseTree shows a more significant performance loss as the expertise context increases in size. These larger sizes are more likely to result in the observation of false correlations, inducing unnecessary splits which the Incremental ExpertiseTree is unable to revert.

\subsection{Execution Time}

\begin{figure}
\flushleft
     \includegraphics[height=.15\textwidth]{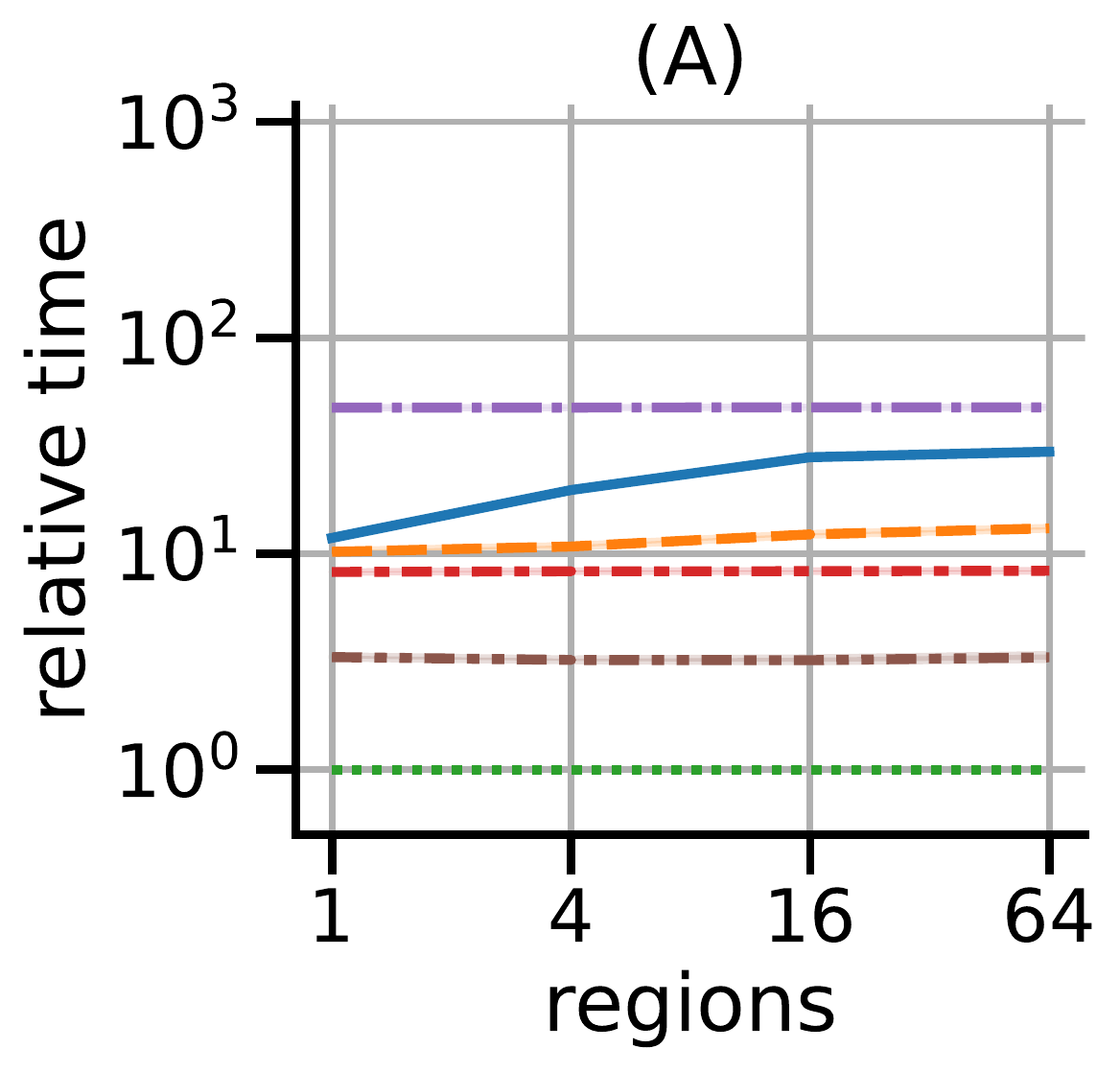}
     \includegraphics[height=.15\textwidth]{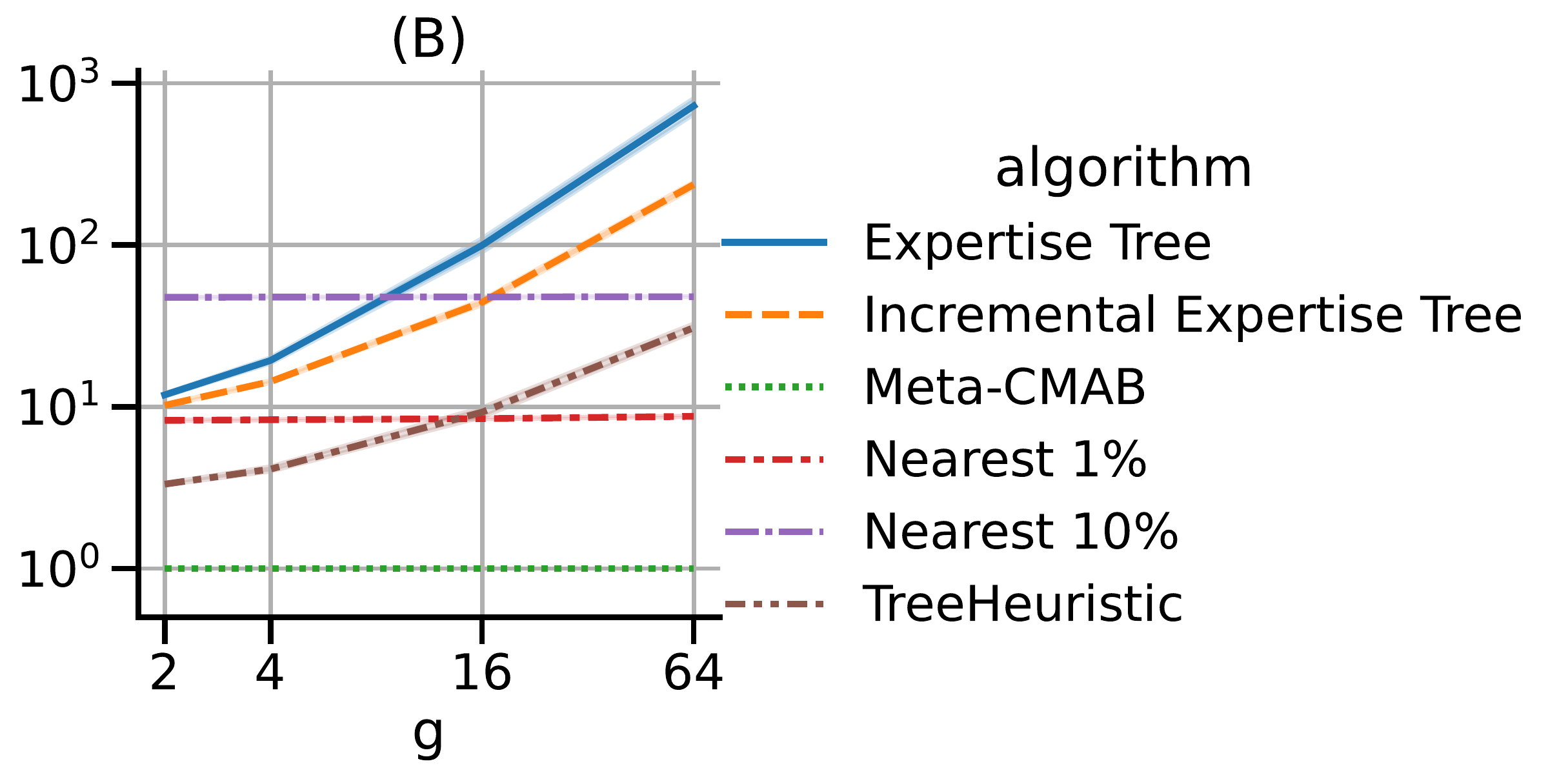}

     \caption{%
     Relative execution time for a varying number of  (A) regions or (B) expertise context size. Relative time is obtained by dividing the execution time of algorithms by the execution time of Meta-CMAB. 
      Both axes are logarithmic. The execution times of the ExpertiseTree methods grow logarithmically with the number of regions, and quasilinearly with the expertise context size.
   }\label{fig:depth_performance}
\end{figure}

Figure \ref{fig:depth_performance} illustrates that the execution time of most algorithms is constant in function of the number of regions or the expertise context size. In particular, the Nearest Neighbor algorithms learn a single model each round, on a set of experiences which is independent of the number of regions or the expertise context size. The neighborhood size does have a significant impact however. A larger neighborhood results in models trained on more experiences, hence a higher execution time. Note that a larger expertise context size increases the cost of distance calculations, resulting in slightly increased execution times.

The trade-off for ExpertiseTree's improvements in terms of average reward is a significant increase in computation time. These plots confirm the theoretical increases in execution time, with a logarithmic dependence on the number of regions and a linear dependence on the size of the expertise context. The resulting execution time can be prohibitive for time-sensitive problems but remains reasonable for, for example, problems with human expertise. Note that in particular, the cost of the ExpertiseTree results from the number of potential splits, thus from the size of the expertise context. This can be improved by dismissing uncorrelated features (either preemptively or based on the observed data).

\section{Conclusion}
In this work, we tackled a generalization of bandits with expert advice which explicitly models localized expertise. Methods which account for localized expertise are essential in real-world settings in which human and artificial expertise depends on a set of features. Observing that prior algorithms in this setting were not conditioned on this set of features, we proposed two novel algorithms of increasing performance and complexity. The first algorithm relies on nearest-neighbor queries to identify relevant experiences but fails when the size of the neighborhood is inappropriate, or when the distance measure is not reflective of similarity in expert quality. To address these drawbacks, our expertise tree approach learns a tree which splits the expertise space in function of the relevant features and then learns to act on all expert advice in each learned partition. This allows it to perform strongly even when the number of regions or the size of the expertise context is large. 
To enhance its applicability, we believe future work should address ExpertiseTree's high computational cost. In particular, online feature selection could be beneficial to reduce the computational cost induced by high dimensional contexts.

\section*{Acknowledgments} A.A. is supported by a FRIA grant (nr. 5200122F) by the National Fund for Scientific Research (F.N.R.S.) of Belgium. T.L. is supported by the F.N.R.S. project with grant numbers 31257234 and 40007793, the F.W.O. project with grant nr. G.0391.13N, the Service Public de Wallonie Recherche under grant n\textdegree 2010235–ARIAC by DigitalWallonia4.ai. T.L and A.N. benefit from the support of the Flemish Government through the AI Research Program. T.L., V.T and A.N. acknowledge the support by TAILOR, a project funded by EU Horizon 2020
research and innovation program under GA No 952215. The resources and services used in this work were provided by the VSC (Flemish Supercomputer Center), funded by the Research Foundation - Flanders (FWO) and the Flemish Government.

\bibliographystyle{icml2023}
\bibliography{references}

\newpage
\begin{appendices}
\addappheadtotoc

\section{Dataset Selection}

We sample datasets from the openmpl \citep{vanschoren2014openml} repository which match the following criteria.

\begin{enumerate}
    \item Sufficient number of instances: in order to allow us to repeat our experiments on each dataset, we restrict our experiments to datasets which contain at least $10,000$ instances.
    \item Classification: we restrict our experiments to classification bandits.
    \item Missing Values: we only consider complete datasets and thus drop datasets which contain missing values. 
    
\end{enumerate}

These filters result in the selection of the following datasets:
mushroom,
    adult,
    letter,
    nursery,
    pendigits,
    BNG(page-blocks,nominal,295245),
    BNG(glass,nominal,137781),
    BNG(tic-tac-toe),
    BNG(vote),
    electricity,
    covertype,
    kropt,
    BNG(breast-w),
    BNG(page-blocks),
    BNG(glass),
    mammography,
    eye\_movements,
    mozilla4,
    KDDCup99,
    MagicTelescope,
    Click\_prediction\_small,
    artificial-characters,
    bank-marketing,
    eeg-eye-state,
    kr-vs-k,
    ldpa,
    skin-segmentation,
    spoken-arabic-digit,
    walking-activity,
    volcanoes-b1,
    creditcard,
    Amazon\_employee\_access,
    CreditCardSubset,
    PhishingWebsites,
    Diabetes130US,
    numerai28.6,
    fars,
    shuttle,
    Run\_or\_walk\_information,
    tamilnadu-electricity,
    jungle\_chess\_2pcs\_raw\_endgame\_complete,
    MiniBooNE,
    jannis,
    helena,
    microaggregation2

For each of these datasets we dummy and normalize all features. 

\end{appendices}
\end{document}